\documentclass[letterpaper, 10 pt, conference]{ieeeconf}
\usepackage[utf8]{inputenc}
\usepackage{array}
\usepackage{float}
\usepackage{amsmath}
\usepackage{graphicx}
\usepackage{caption}
\usepackage{subcaption}
\usepackage{wrapfig}

\let\labelindent\relax
\usepackage{enumitem}
\makeatletter

\providecommand{\tabularnewline}{\\}

\usepackage{authblk}
\usepackage[ruled,vlined]{algorithm2e}
\usepackage{multicol}
\usepackage{inputenc}
\usepackage{amsfonts}
\usepackage{amsmath}
\usepackage{amssymb}
\usepackage[T1]{fontenc}
\usepackage{microtype}
\usepackage[dvipsnames]{xcolor}
\usepackage{graphicx}
\usepackage{graphics}
\usepackage[font=footnotesize]{caption}
\usepackage{subcaption}
\usepackage{booktabs}
\usepackage{multirow}
\usepackage{cite}
\usepackage[export]{adjustbox}
\usepackage{breqn}
\usepackage{acronym}
\usepackage[keeplastbox]{flushend}
\usepackage{setspace}
\usepackage{bm}
\usepackage{stackengine}

\usepackage{smartdiagram}
\usepackage{tikz}
\usesmartdiagramlibrary{additions}

\usepackage{hyperref}
\definecolor{azure}     {rgb}{0,0.5,1}
\definecolor{dkpowder}  {rgb}{0,0.2,0.7}
\definecolor{deepred}   {rgb}{0.7,0,0}
\definecolor{deepblue}  {rgb}{0,0,0.7}
\definecolor{deepgreen} {rgb}{0,0.5,0}
\definecolor{deeporange}{rgb}{0.91, 0.41, 0.17}
\hypersetup{%
	pdfpagemode  = {UseOutlines},
	bookmarksopen,
	pdfstartview = {FitH},
	colorlinks,
	linkcolor = {dkpowder},
	citecolor = {dkpowder},
	urlcolor  = {dkpowder},
}

\usepackage{listings}

\lstdefinelanguage{BibTeX}{%
keywords={%
    @article,@book,@collectedbook,@conference,@electronic,@ieeetranbstctl,%
    @inbook,@incollectedbook,@incollection,@injournal,@inproceedings,%
    @manual,@mastersthesis,@misc,@patent,@periodical,@phdthesis,@preamble,%
    @proceedings,@standard,@string,@techreport,@unpublished%
},
comment=[l][\itshape]{@comment},
sensitive=false,
}

\renewcommand{\thetable}{\arabic{table}}

\DeclareMathOperator*{\argmax}{\arg\!\max}

\graphicspath{{/images}}
\makeatletter
\def\endthebibliography{%
	\def\@noitemerr{\@latex@warning{Empty `thebibliography' environment}}%
	\endlist}
\makeatother

\definecolor{label-running} {RGB}{ 31,119,180}
\definecolor{label-walking} {RGB}{255,127, 14}
\definecolor{label-jumping} {RGB}{ 44,160, 44}
\definecolor{label-standing}{RGB}{148,103,189}
\definecolor{label-sitting} {RGB}{140, 86, 75}
\definecolor{label-lying}   {RGB}{127,127,127}
\definecolor{label-falling} {RGB}{188,189, 34}
\definecolor{label-transit} {RGB}{ 23,190,207}

\IEEEoverridecommandlockouts

\title{\LARGE \bf Deep Residual Reinforcement Learning based \\ Autonomous Blimp Control}
\author{Yu Tang Liu$^{1,2}$, Eric Price$^{2,1}$, Michael J. Black$^{1}$ and Aamir Ahmad$^{2,1}$ \thanks{$^1$Max Planck Institute for Intelligent Systems, 72076 T{\"u}bingen, Germany.
$^2$Institute for Flight Mechanics and Controls, The Faculty of Aerospace Engineering and Geodesy, University of Stuttgart, 70569 Stuttgart, Germany. (yutang.liu, eric.price, black)@tuebingen.mpg.de, aamir.ahmad@ifr.uni-stuttgart.de}
}

\makeatother

\begin{document}
\maketitle 

\makeatletter


\makeatother
\begin{abstract}

Blimps are well suited to perform long-duration aerial tasks as they are energy efficient, relatively silent and safe. 
\
To address the blimp navigation and control task, in previous work we developed a hardware and software-in-the-loop framework and a PID-based controller for large blimps in the presence of wind disturbance.
\
However, blimps have a deformable structure and their dynamics are inherently non-linear and time-delayed, making PID controllers difficult to tune. Thus, often resulting in large tracking errors.
\
Moreover, the buoyancy of a blimp is constantly changing due to variations in ambient temperature and pressure. 
\
To address these issues, in this paper we present a learning-based framework based on deep residual reinforcement learning (DRRL), for the blimp control task. 
\
Within this framework, we first employ a PID controller to provide baseline performance. Subsequently, the DRRL agent learns to modify the PID decisions by interaction with the environment. We demonstrate in simulation that DRRL agent consistently improves the PID performance. Through rigorous simulation experiments, we show that the agent is robust to changes in wind speed and buoyancy. In real-world experiments, we demonstrate that the agent, trained only in simulation, is sufficiently robust to control an actual blimp in windy conditions. We openly provide the source code of our approach at \url{https://github.com/robot-perception-group/AutonomousBlimpDRL}.
\end{abstract}



\section{Introduction}
\label{sec:Introduction}


Autonomous unmanned aerial vehicles (UAVs) are becoming increasingly popular for various tasks, such as search and rescue, payload (medicine, food) delivery in difficult-to-reach areas, aerial cinematography and wildlife monitoring \cite{gonzalez2016unmanned}. Current solutions rely on quadcopters and fixed-wings. Although quadcopters can hover in a fixed position, they are not able to accomplish long-term missions due to their short battery life. The situation is opposite for the fixed-wings, which requires moving constantly to stay airborne. Therefore, for tasks involving long flight times, carrying more payload and hovering over a small region, the use of autonomous blimps is an attractive solution. However, autonomous blimp control remains a challenging problem, which we address in this paper using a learning based approach.

Classic blimp controller design usually relies on PID controllers \cite{1013654,takaya2006pid} and nonlinear control \cite{1570450,LIU2020105610}. PID struggles with plant nonlinearity, and nonlinear control methods require a dynamic model of the system which is often difficult to acquire (i.e. friction, wind, aerodynamic effect, etc.). Deep reinforcement learning (DRL), on the other hand, is a new control framework that has achieved success in a variety of applications that present similar challenges \cite{NIPS2003_2455, bellemare2020autonomous}.

For blimp control, the knowledge of some physical parameters, such as friction, aerodynamic effect, etc., are not negligible but remain difficult to estimate. A model-free DRL approach is particularly useful in such a case as it allows an agent to learn a control policy without any pre-specified physics and without the need to estimate those parameters explicitly.
\
However, training such a model-free DRL agent requires significant amount of data and computational resources. The trained agent could also often learn unexpected and unsafe maneuvers as it only exploits the reward function. For example, an autonomous blimp agent can learn to fly backwards and still receive a high reward, while in reality such behavior is undesired as it can damage the hardware. Our insight to address these two issues is to leverage a classical model-free approach, e.g., PID, to constraint the policy search space. We do this by developing a novel framework based on deep residual RL (DRRL) \cite{silver2018residual} that combines the advantages of both classical control and reinforcement learning. The use of the classical method in this framework not only provides stability in training but also implicitly outlines a safe behavior for the agent, constraining the policy search space and avoiding undesired behaviors.

\begin{figure}[t]
	\centering
	\includegraphics[width=0.5\textwidth]{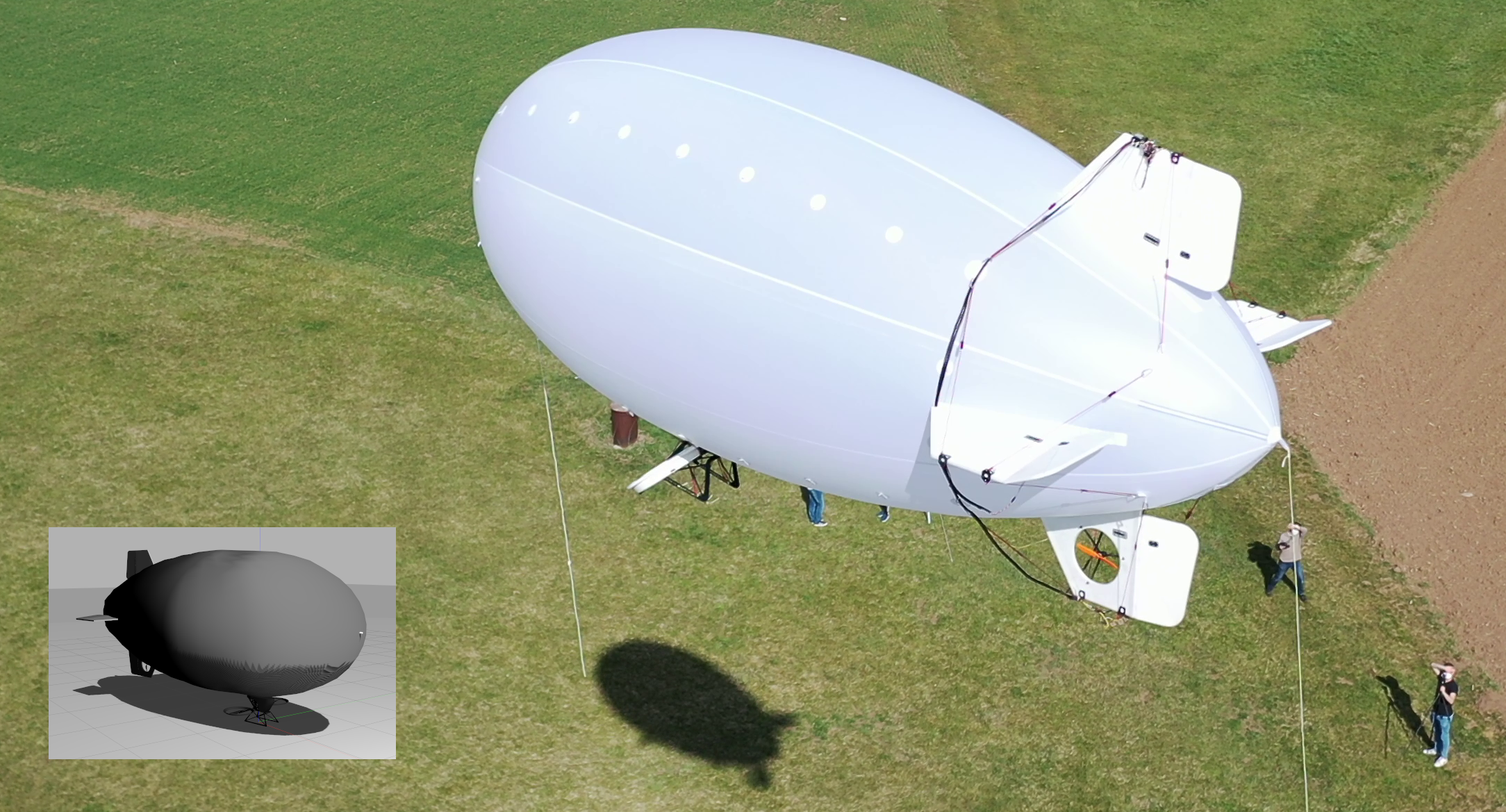}
	\caption{Our autonomous blimp during a flight. Unlike common designs, our blimp has thrust vectoring which increases its agility. Inset: its gazebo model.}
	\label{fig:blimp_body}
\end{figure}


The training process can also be unstable due to the partial observable nature of the environment, e.g. wind and buoyancy, and this effect is exacerbated by the time-delayed blimp dynamics. To address this issue, we integrate an LSTM (long short term memory)\cite{hochreiter1997long} layer in our policy model to reduce the effect of partial observability.

Nevertheless, our DRRL framework still requires substantial training experience to derive a working RL policy. We address this problem by training the agent in a software-in-the-loop (SITL) simulation setup \cite{price2020simulation} and parallelizing it to accelerate this process.

Furthermore, to deploy the agent on the real blimp, it is necessary to i) address the issue of sim-to-real gap, and ii)  maintain smoothness in the actuator commands. Thus, in simulation, we apply domain randomization during training to improve the robustness of the agent. To protect the actuator and reduce the effects of chattering, we only include the increment of the actuator command instead of the actuator command itself (e.g., rotor acceleration instead of rotor speed) in the action space of the agent. 

In summary, the novel contribution of this paper is a model-free DRRL-based approach for autonomously controlling a large blimp in forward velocity, yaw and altitude, simultaneously, in outdoor moderate wind conditions. Through rigorous simulations we show that our method outperforms state-of-the-art approaches based on a PID and is robust to different flight contexts, e.g., changes in wind speed and buoyancy. Finally, through real-world experiments, we demonstrate that using our approach we obtain a robust control policy that seamlessly generalizes to the real blimp.
 

\section{Related work}
\label{sec:2_related}

Control methods for blimps and airships, which have similar control schemes, have been well studied \cite{5420403}. Classic approaches usually rely on \textbf{PID controllers}\cite{1013654, 770044, 894672, takaya2006pid}. While being simple and robust, they often suffer from plant nonlinearity. To overcome this, advanced approaches have been developed using \textbf{nonlinear control theory}, such as inverse optimal tracking control\cite{1554402}, dynamic inversion control\cite{1570450}, backstepping control\cite{4470489}, robust control\cite{Cheng2018}, and model predictive control \cite{4776979}. However, optimal control usually requires an accurate dynamic model which can be difficult to acquire, while robust control handles parameter uncertainty by trading-off the performance. Another key drawback is the lack of any real-world experiments and validation in most of these works. The buoyancy of a real blimp can change significantly due to the fluctuations in temperature over short time periods. The weight distribution could also vary and thus reduce the altitude control performance. Unfortunately, these effects have not been addressed in any of the prior works so far.

On the other hand, recently there has been a surge of interest in applying RL to robotics \cite{singh2021reinforcement}. The earliest works include Gaussian processes (GPs) for system identification of a blimp \cite{4209179} and its combination with value iteration and Q-learning approaches \cite{5152660, 4399531} for blimp altitude control only. Despite sample efficiency, GPs are hard to scale up with problem dimensions and demand higher computational resources. As a result, they are able to achieve success only on low dimensional tasks, such as 1-D altitude control, whereas in our approach we show that the agent can feasibly and successfully learn a 3-dimensional task (forward velocity, yaw and altitude control). \textbf{DRL}, on the other hand, leverages deep neural networks (NNs) for policy approximation. Thus, its policy class can be used for higher dimensional tasks. For example, authors in \cite{nie2019three} train two DQN agents for rudder and elevator control of a blimp, respectively, and demonstrate better performance than a PID controller in simulation. 
\
The main challenge with DRL, however, is the lack of sample efficiency. In order to scale up the DRL formulation with the problem dimension, a highly increased amount of environment interactions is needed by the agent. Other challenges include, but are not limited to, adapting a trained policy to real-world scenarios \cite{zhang2019bridging} and action smoothness\cite{caps2021}. Furthermore, issues such as partial observability, disturbances and noise could also lead to unexpected behaviors. As described in the introduction, in our approach we address all these issues through our novel \textbf{DRRL}-based framework, training parallelization, domain randomization and action space design.





\section{Methodology}

\label{sec:Methodology}

In this section, we first describe the blimp and the MDP problem formulation. Then we introduce the main goal of the work, the \textit{blimp control task} (Sec.\ref{subsec:Blimp-Control-Task}). The objective in the \textit{blimp control task} is to navigate the blimp to any given waypoint within the space $L^{3}$m$^{3}$, where $L$ is the dimension of the bounding box. This is followed by our novel DRRL-based framework that describes our approach for the \textit{blimp control task}. Subsequently, we describe \textit{yaw control task} (Sec.\ref{subsec:Yaw-Control-Task}), where the aim is to control the blimp to a desired yaw angle with the tail rotor. The aim of this simplified task is to perform ablation study.

\subsection{Preliminaries}
\label{subsec:Preliminary}

We first briefly describe our blimp (complete details are in our previous work \cite{price2020simulation}), which has 8 actuators. The two main motors (thrusters), $m_{1,2}$, are
attached to a servo, $n_{0}$, which allows thrust vectoring. At the
tail of the blimp, four fins, $f_{0:3}$, two positioned vertically and two horizontally, control yaw and pitch angle, respectively.
There is a tail motor, $m_{0}$, attached to the lower vertical fin, generating
horizontal thrust allowing further yaw controllability. Therefore, the state
vector of the actuators can be denoted as

\begin{equation}
s_{t}^{act}=(m_{(0:2)},n_0,f_{(0:3)})_{t}\in\mathbb{R}^{8},\label{eq:actuator_state}
\end{equation}

\subsection{Markov Decision Process}
\label{subsec:Markov-Decision-Process}

We consider the RL problem as an infinite horizon discrete time Markov
Decision Process, $M$, defined by a tuple $(S,A,P,R,\gamma)$ \cite{sutton2018reinforcement}.
At any time step $t\in\mathbb{R}^{+}$ and state $s_{t}\in\mathbb{R}^{S}$,
an agent draws an action $a_{t}$ from a continuous action space $a\in\mathbb{R}^{A}$
given the policy distribution $a_{t}\sim\pi_{\theta}(\cdot|s_{t})$
parameterized by $\theta$. The environment then samples the next
state from an unknown transition distribution, i.e. $s_{t+1}\sim P(\cdot|s_{t},a_{t})$.
A reward is received based on some reward function $r_{t}=R(s_{t},a_{t})$.
Given the discount factor $\gamma\in[0,1)$, the goal of the agent
is to obtain the optimal policy parameter $\theta$ that maximizes the expected value of the cumulative discounted reward (\ref{eqn:total_sum_of_discounted_reward}),

\begin{equation}
\begin{small}
	\begin{gathered}\pi^{*}=\argmax_{\pi_{\theta}}\mathop{\mathbb{E}}_{\pi}\left[\sum_{t}^{\infty}\gamma^{t}r_{t}|a_{t}\sim\pi(\cdot|s_{t}),s_{t+1}\sim P(\cdot|s_{t},a_{t})\right]\end{gathered}
	\label{eqn:total_sum_of_discounted_reward}
\end{small}
\end{equation}


\subsection{Blimp Control Task}
\label{subsec:Blimp-Control-Task}
We formulate the problem as a path following task as seen in previous
works \cite{1626776,nie2019three,5611169}. In this setting, an imaginary
path reference is generated based on waypoints for the controller
to follow. Casting the path following task as a DRL problem, in this
section we derive the observation space and action space representation. 

Since the blimp does not have a lateral
movement control, we only consider longitudinal, altitude, and velocity
control. This allows us to easily decompose the problem into planar, altitude, and velocity control. The objective
of the planar control is to control the blimp to arrive
at any waypoint in the xy-plane, the altitude control is to
reach the desired z, and the velocity control is to track the desired velocity. 

\begin{figure}[H]
	\begin{centering}
		\includegraphics[width=0.8\columnwidth]{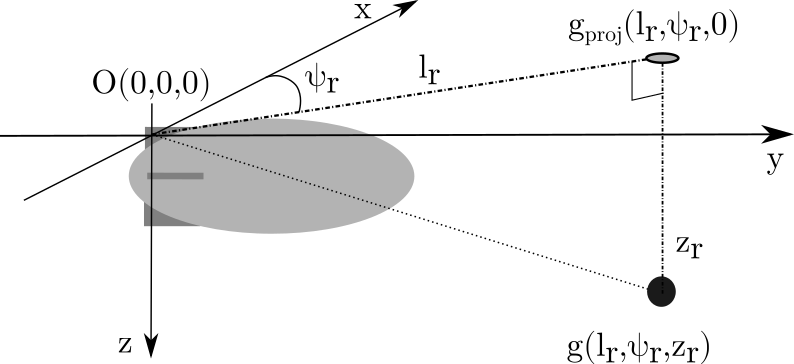} 
		\par\end{centering}
	\caption{The origin indicate the position of GPS sensor in NED coordinate. \label{fig: blimp_diagram}}
\end{figure}

Given the blimp position at $\overline{O}=(0,0,0)$ and velocity $v_o\in \mathbb{R}$,
a target waypoint at
$\overline{g}=(l_{r},\psi_{r},z_{r})$ in body frame cylindrical coordinates with desired velocity $v_g\in \mathbb{R}$
(Fig.\textasciitilde\ref{fig: blimp_diagram}), the control objective
of the planar control is the minimization of the relevant
distance and yaw angle, i.e. $\min_{a\in A}(|l_{r}|,|\psi_{r}|)$. The
objective of the altitude and velocity control is to minimize the relevant altitude and the relevant velocity, respectively, i.e., $\min_{a\in A}(|z_{r}|)$, $\min_{a\in A}(|v_r|)$, where $v_r=v_g-v_o$.

We denote the velocity vector of the blimp as $\overline{V}=(u,v,w)\in \mathbb{R}^3$, and attitude
(roll, pitch, yaw) as $\overline{\Phi}=(\phi,\theta,\psi)\in \mathbb{R}^3$. Assuming near zero
lateral movement in the blimp (i.e. $v,\phi\simeq0$), the velocity
and pitch angle can be encoded by velocity magnitude ($v_o=||(u,v,w)||_{2}$)
and the altitude velocity ($w=v_o \sin\theta$), alone. Therefore, the base state vector is
\begin{equation}
s_{t}^{blimp}=(l_{r},\psi_{r},z_{r},v_{r}, v_{o}, w)_{t}\label{eq:state_vector}.
\end{equation}

We augmented the base state vector with additional components, based on the insights as explained below. It was observed that the training progress becomes more stable if yaw velocity, $\omega_{\psi}$ augmented to the base state vector. The airspeed sensor readings, $v_{air}\in \mathbb{R}$, were augmented to enhance robustness against the wind. To prevent overshoot when reaching a waypoint in the planar control task, we augmented $\psi_{r}'$, the relative yaw angle of the blimp with respect to the subsequent waypoint. Consequently, the extended state representation is 
\begin{equation}
s_{t}^{blimp'}=(s^{blimp},\omega_{\psi},v_{air},\psi_{r}')_{t}\label{eq:informative_state_vector}.
\end{equation}

\subsection{Novel DRRL-based framework}
\begin{figure}[t]
	\centering
	\includegraphics[width=0.4\textwidth]{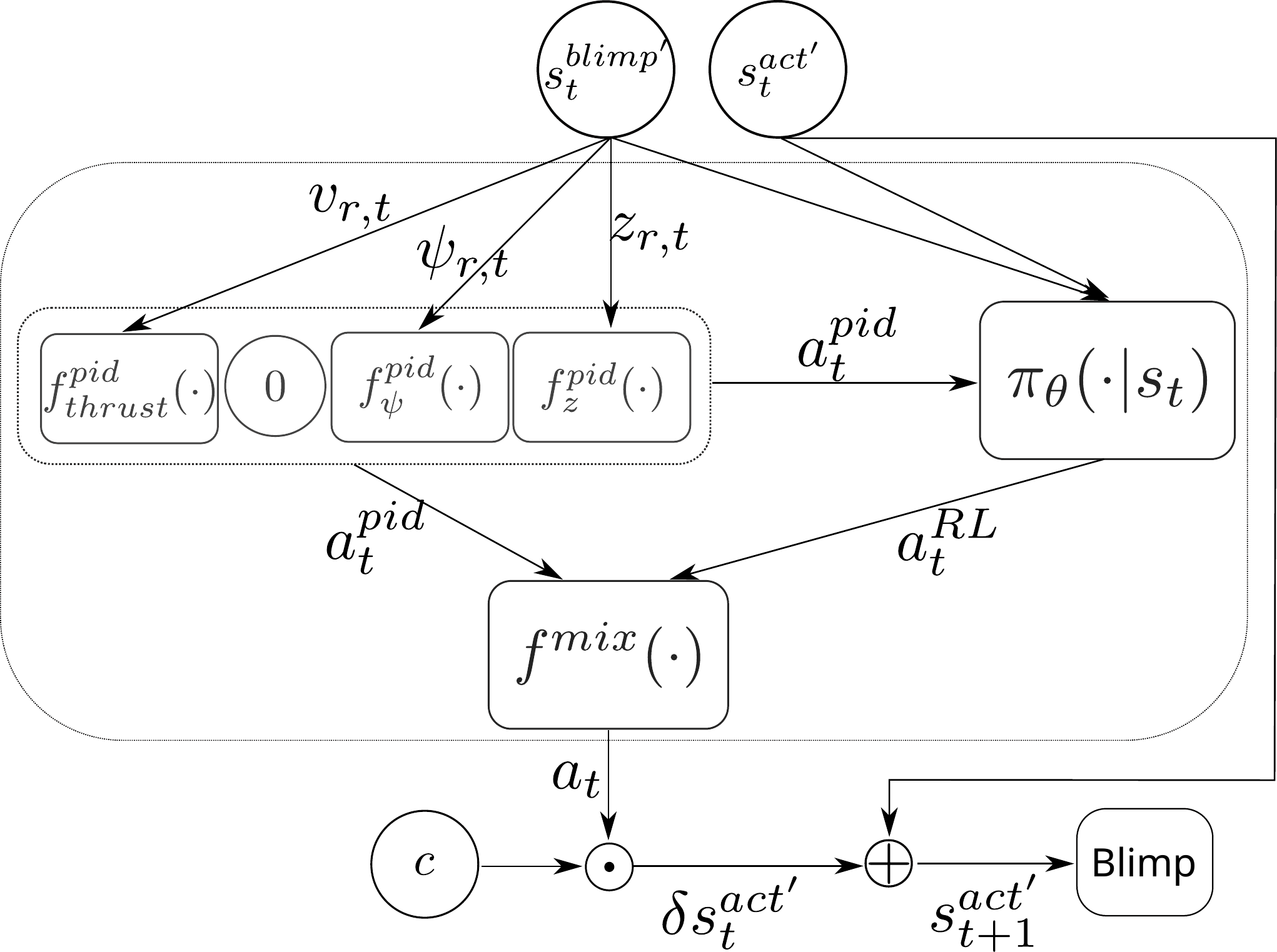}
	\caption{DRRL control diagram. Note that input and output of the policy network and PID controllers are scaled and clipped to the range $(-1,1)$ }
	\label{fig:drrl_architecture}
\end{figure}

Our DRRL framework consists of two controllers -- a stability provider and a performance optimizer, respectively. The classical approach offers stability guarantees and basic tracking performance which is the role usually played by a PID controller or a robust controller to enlarge region of attraction. Performance optimizer within this framework is a DRL agent that can learn to adjust the control decisions of the stability provider in order to maximize its own reward function. The control command from these two controllers are then mixed by a mixer $f^{mix}(\cdot)$, which we will described later. The overall structure is displayed in Fig.~\ref{fig:drrl_architecture}. In this paper, we choose PID controller as our stability provider for its simplicity and robustness. It integrates well with the DRL agent as it is also a model-free method. No dynamic model is required with this combination. Though its performance degrades quickly outside the tuned speed range, the system nevertheless remains stable and can still bring the blimp closer to the waypoint. 

\subsubsection{PID Controller}
\label{subsec:PID-Controller}

The PID command, $a_{t}^{pid}\in\mathbb{R}^{4}$ is determined as follows (\ref{eq:PID_action}),
\begin{equation}
	\begin{array}{ll}
		a_{t}^{pid}&=[a_{thrust}, a_{servo}, a_{\psi}, a_{z}]_{t}^{pid}\\
				   &=[f^{pid}_{thrust}(v_{r,t}), 0, f^{pid}_{\psi}(\psi_{r,t}), f^{pid}_{z}(z_{r,t})]
	\end{array}\label{eq:PID_action}
\end{equation}

where $f^{pid}(x)=k_{p}x+k_{i}\int x+k_{d}\dot{x}$. Since it is difficult to design a PID-based servo control, we leave this completely for the DRL agent to control.

\subsubsection{Observation and Action Space for the DRL agent}
\label{subsec:Observation-and-Action-Space}

The full actuator state, $s_{t}^{act}$, is described in (\ref{eq:actuator_state}).
Since we forbid differential thrust, symmetric actuators are
always in the same state. Thus, we feedback only one of them
(i.e. $m_{1}=m_{2},f_{0}=f_{1},f_{2}=f_{3}$). The tail motor is controlled
and observed together with bottom fin (i.e. $m_{0}=f_{2}$). The reduced
state of actuators is therefore defined as $s_{t}^{act'}=(m_{1},n_0,f_{(0,2)})_{t}\in\mathbb{R}^{4}$.
The full state $s_{t}$ for the DRL formulation, as used in (\ref{eqn:total_sum_of_discounted_reward}),
is now obtained below as the concatenation of $s_{t}^{blimp'}$, $s_{t}^{act'}$, and $a_{t}^{pid}$

\begin{equation}
\begin{gathered}[l]
	s_{t}=(s_{t}^{blimp'},s_{t}^{act'}, a_{t}^{pid})\end{gathered}
\label{eq:full_state_representation}
\end{equation}

Note that all states are scaled to the range $[-1,1]$ and zero-initialized.
The RL command, $a_{t}^{RL}\in\mathbb{R}^{4}$, is chosen based
on the state vectors. Then the joint action command, $a_{t}\in\mathbb{R}^{4}$,
is simply the mixture of RL and PID actions.

\begin{equation}
\begin{gathered}a_{t}^{RL}\sim\pi(\cdot|s_{t})\\
a_{t}=f^{mix}(a_{t}^{pid},a_{t}^{RL})
\end{gathered}
\label{eq:joint_action}
\end{equation}

We introduce 3 types of mixer: absolute, relative, and hybrid mixer
(\ref{eq:absolute_mix}-\ref{eq:hybrid_mix}). 
Absolute mixer offers RL agent more authority and is expected to have the highest performance after convergence at the cost of performance drop during exploration. This property is reversed for the relative mixer. 
Since the absolute mixer is too aggressive and requires rigorous tuning of the beta parameter,  whereas the relative mixer is too conservative to change the system's inherent stability properties, we introduce a hybrid mix as an intermediate solution. 
\begin{align}
& f^{mix}_{abs}(x,y)=(1-\beta)x+\beta y\label{eq:absolute_mix}\\
& f^{mix}_{rel}(x,y)=x(1+\beta y)\label{eq:relative_mix}\\
& f^{mix}_{hyb}(x,y)=(1-\alpha)f^{mix}_{abs}(x,y)+\alpha f^{mix}_{rel}(x,y)\label{eq:hybrid_mix},
\end{align}where $(\alpha,\beta)\in[0.0,1.0]$. To reduce the effect of chattering in the actuator state, we avoid mapping joint command, $a_{t}$, to the actuator state directly. Instead, it is first mapped to the increment of actuator state $\delta s_{t}^{act'}\in\mathbb{R}^{4}$ by element-wise multiplication with a constant vector, $c\in\mathbb{R}^4$,
and then update the actuator state. The process is described below in (\ref{eqn:actuator_state_update}). This way, we prohibit sudden significant changes in actuator states. Since our electronic speed control filters small changes, the damage from chattering effect is almost diminished.  The disadvantage of this approach is that the control agility is reduced due to an additional pole introduced at the origin.

\begin{equation}
\begin{gathered}
	\delta s_{t}^{act'}=c\odot a_{t}\\
	s_{t+1}^{act'}\leftarrow s_{t}^{act'}+\delta s_{t}^{act'}
\end{gathered}
\label{eqn:actuator_state_update}
\end{equation}

We summerized the overall control architecture in Fig.~\ref{fig:drrl_architecture}.

\subsubsection{Reward Function}
\label{subsec:Reward-Function}

The navigation requires moving the robot in space by specifying a
target position or following a sequence of waypoints. The reward function
is defined by (\ref{eqn:reward_function})

\begin{equation}
\begin{gathered}[l]r_{t}=w_{0}r_{t}^{success}+w_{1}r_{t}^{track}+w_{2}r_{t}^{act}+w_{3}r_{t}^{bonus},\end{gathered}
\label{eqn:reward_function}
\end{equation}where $w_{0:3}\in\mathbb{R}$. The agent receives a
success reward, $r_{t}^{success}$, if the task is completed, i.e., a waypoint is successfully reached within a certain threshold $\epsilon$. Tracking
reward, $r_{t}^{track}$, indicates the tracking performance as defined in (\ref{eqn:individual_reward}). Action reward, $r_{t}^{act}$, is defined to regularize actuator commands. Bonus reward, $r_{t}^{bonus}$, specifies additional desired control property of preventing overshoot.

\begin{equation}
\begin{array}{l}
r_{t}^{success}=\begin{cases}
1 & \text{if \ensuremath{d(s_{blimp},s_{target})\leq\epsilon} }\\
0 & \text{otherwise}
\end{cases}\\
	r_{t}^{track}=-i_{0}|z_{r}|-i_{1}|l_{r}|-i_{2}|\psi_{r}|-i_{3}|v_{r}|,\\
	r_{t}^{act}=-j_{0}|m_{0}|-j_{1}|m_{1}|-j_{1}|m_{2}|,\\
	r_{t}^{bonus}=-k_{0}|\psi_r'|/(1+l_r),
\end{array}\label{eqn:individual_reward}
\end{equation}where $d(s_{blimp},s_{target})$ measures euclidean distance between
the blimp and the target waypoint position's 2D projections on the ground plane. Parameters, $w_{0:3}, i_{0:3}, k_{0} \in\mathbb{R}$, are derived via manual tuning and $j_{0:1}\in\mathbb{R}$ approximate the energy consumption of the rotors. The bonus reward is designed to reduce overshoot by reducing the relative yaw angle to the next waypoint when the blimp approaches current target waypoint. 

\subsection{Yaw Control Task}
\label{subsec:Yaw-Control-Task}

Here, the agent observes yaw-related states and outputs a yaw command to control the tail motor, $m_0$, together with a PID controller. The objective is to minimize the relative yaw angle, i.e. $\min_{a\in A}(|\psi_{r}|)$.
Concretely, the observation space is defined as $s_t^{yaw}=(\psi_r,\omega_{\psi},a_{\psi}^{pid})_t$ and the action space $a_t^{yaw}=f^{mix}(\pi(\cdot|s_t^{yaw}), a_{\psi,t}^{pid})$ where $s_t^{yaw}\in\mathbb{R}^3$ and $a_t^{yaw}\in\mathbb{R}$. The actuator state update is described as in (\ref{eqn:actuator_state_update}) with $c\in\mathbb{R}$. Different from (\ref{eqn:reward_function}), the reward function in this task, only combines success reward and tracking reward, or $r_t^{yaw}=w_0r_t^{success}+w_1r_t^{track}$. The task is considered as success if the agent's relative yaw angle is kept small for a certain period of time, as described in (\ref{eq:yaw_control_reward}).

\begin{equation}
	\begin{array}{l}
		r_{t}^{success}=\begin{cases}
			1 & \text{if \ensuremath{T(|\psi_r|\leq\epsilon)\geq T_{success}} }\\
			0 & \text{otherwise}
		\end{cases}\\
		r_{t}^{track}=-|\psi_{r}|,
	\end{array}\label{eq:yaw_control_reward}
\end{equation}where $T(\cdot)$ is a time counting function and $T_{success}\in\mathbb{R}^+$.


\subsection{Training Setup}

\label{subsec:Training-Setup}

In this section, we describe the important factors attributed
to the robustness of the trained policy. At the beginning of each training episode, several waypoints
are sampled in the space based on the position of previous waypoint. When the blimp
reaches the current waypoint, it receives a success reward and activates the next waypoint. During training, observations and actions are injected with noise. Lastly, we apply domain randomization and sample new environment variable for
each episode according to Table.\ref{tab:domain-randomization}.

\begin{table}[h]
	\centering
	\resizebox{0.3\textwidth}{!}{%
	\begin{tabular}{|c|>{\centering}p{0.3\columnwidth}|}
	\hline 
	variable  & range\tabularnewline
	\hline 
	\hline 
	wind in xy direction $(m/s)$  & {[}-1.5, 1.5{]}\tabularnewline
	\hline 
	wind in z direction $(m/s)$ & {[}-0.15, 0.15{]}\tabularnewline
	\hline 
	buoyancy $(100\%)$ & {[}0.9, 1.1{]}\tabularnewline
	\hline 
	freeflop angle $(\cdot)$ & {[}0.0, 1.5{]}\tabularnewline
	\hline 
	collapse $(\cdot)$ & {[}0.0, 0.02{]}\tabularnewline
	\hline 
	deflation rate $(\cdot)$ & {[}0.0, 1.5{]}\tabularnewline
	\hline 
	\end{tabular}}
	\caption{Freeflop angle, collapse, and deflation rate determine maximum fin decline angle, the stiffness of the cable that holds the fin, and the rigidity of the blimp hull respectively, which jointly affect fin decline angle and reduce the lift and drag force\cite{price2020simulation}. 
	\label{tab:domain-randomization}}
\end{table}

\begin{figure}[b]
	\centering
	\includegraphics[width=0.25\textwidth]{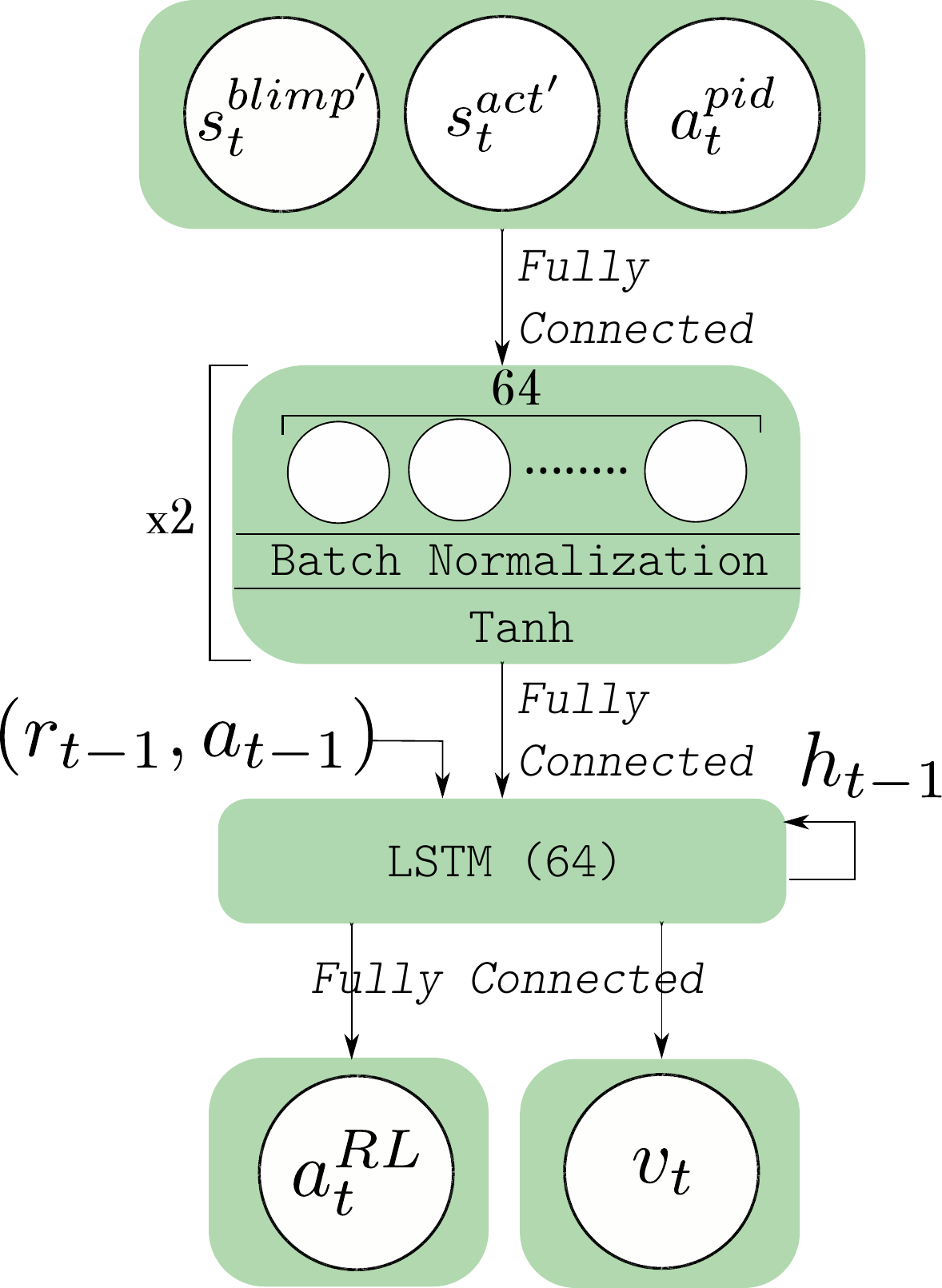}
	\caption{Following \cite{andrychowicz2020matters}, we initialize the output layer with small weights (e.g. $1e^{-12}$), apply \textit{tanh} activation function, and integrate normalization layers to stabilize training.   $h_{t-1}$: hidden state. $r_{t-1}$: previous reward. $a_{t-1}^{RL}$:previous action. $v_t$: estimated value from the critic.}
	\label{fig:nn_connect}
\end{figure}

We train the policy network with PPO, which has achieved strong benchmark performance and training stability. 
To accelerate training, we parallelize the simulation and raise the speed up to 14 folds of the real-time. The architecture
of our policy network (Fig. \ref{fig:nn_connect}) includes an LSTM layer to reduce the impact of partial observability (e.g. wind, bouyancy, etc.). 

\section{Experiment Design and Setup}

\label{sec:4_experiment}

With the real world scenario in mind, we address the following questions through our experiments. Does LSTM architecture reduce the impact of partial observability? What are the properties of the mixers in the DRRL framework? Can the DRRL agent improve the PID controller? How does the agent perform under the presence of disturbance, noise, and parameter uncertainty?  

\subsection{Experimental Setup \label{subsec:Experiment-Setup} and compared methods}

We integrate our DRRL training environment in the ROS/Gazebo SITL simulation
following the OpenAI-Gym framework. The PPO implementation is based
on RLlib. The agent is trained on
a single computer (AMD Ryzen Threadripper 3960X, 24x 3.8GHz, NVIDIA
GeForce RTX 2080 Ti, 11GB). Our simulated blimp model is designed based on our real robotic blimp (see Fig.~\ref{fig:blimp_body}). The following methods are evaluated and compared to each other.
\begin{itemize}[leftmargin=*]
 \item \textbf{DRRL agent}: our proposed approach.
 \item \textbf{PID}: the PID described in Sec.~\ref{subsec:PID-Controller}
 \item \textbf{Baseline}: the baseline is a cascade PID controller, well-tuned to the simulation environment. Our previous work \cite{price2020simulation} has shown that we could deploy it to the real world without tuning, which implies a reliable quality of the simulation and robustness of such approach. This controller directly controls the actuator to follow the velocity reference from a path planner instead of the waypoints (as used by the above 2 methods) and relies on an extended Kalman filter for state estimation and noise filtering. 
\end{itemize}

\subsection{Task Suite \label{subsec:Task-Suite}}
In this section, we describe the design of the two control tasks that were introduced previously. The \textit{Yaw control task} (\ref{subsec:Yaw-Control-Task}) is a simplified task to evaluate different design options. The goal is to acquire the best possible configuration to then train a near-optimal policy for \textit{blimp control task} (\ref{subsec:Blimp-Control-Task}) within limited amount of time. To ensure reproducibility, the training experiments are conducted with $3$ different seeds. Table \ref{tab:exp_parameter} displays the parameters for both tasks. 

\subsubsection{Yaw Control Task}
We carry out an empirical ablation study on training stability of DRRL agent with different PID controller, different policy, and mixer combination. We first use a PD and then a P-control, which correspond to a good and a poor PID controller, respectively, for this task. PD control has stability guarantee while P control is only marginally stable. 


\subsubsection{Blimp Control Task}
We design the DRRL agent for this task based on the conclusion from the ablation study of the \textit{yaw control task}. Despite the difference in task complexity, the PPO agent hyperperameter remains identical. 
We first examine the training progress of the DRRL agent and compare to the PID controller. The training is performed 3 times with different seeds to ensure the reproducibility. 

Then, we investigate its robustness and characteristics in different wind context w.r.t the PID and the Baseline methods. This comparison is performed on results averaged over 7 runs, each lasting for 30 minutes and for 2 desired trajectories (coil and square). Furthermore, these are subject to random uniformly sampled wind direction. The square trajectory consists of 4 waypoints and has 80 meters between each waypoint. The coil trajectory has 30 meter radius covered by 15 waypoints in total. Consecutive waypoints on it are separated by 45 degrees and 42.4m in their projection on the X-Y plane and by 2m in the Z direction. As the square trajectory has longer edges, it is easier to track it as compared to the coil. The coil trajectory is more challenging due to the shorter inter-waypoint distance. In this case, the blimp has to constantly slow down to control the yaw angle which can cause altitude loss.

We test the trained agent on a real blimp with 40 meters square trajectory. The real blimp has several different properties compared to the simulation, e.g. trim weight difference, buoyancy, maximum thrust etc. Many of these effects are not domain randomized during training. That is to say, it is a new flight context that the DRRL agent has not encountered before and thus it pose a great challenge for generalization. 


\begin{table}[h]
	\centering
	\resizebox{0.4\textwidth}{!}{%
	\begin{tabular}{ |p{2.0cm}||p{3.7cm}|p{2.2cm}|  }
		\hline
		\multicolumn{3}{|c|}{Setup details} \\
		\hline
		Group & Name & Value \\
		\hline
		PPO 
		&   learning rate schedule &  $[1e^{-4}, 5e^{-6}]$ \\
		&   $\gamma$  &  $.999$ \\
		&   $\lambda$  &  $.9$ \\
		&   horizon  &  $800$ \\
		&   batch size  &  $22,400$ \\
		&   mini-batch size  &  $2,048$ \\
		&   sgd iterations  &  $32$ \\
		&   gradient clip  &  $1$ \\
		
		\hline
		Environment 
		& 	simulation frequency [hz] & $30$ \\
		& 	policy frequency [hz] & $10$ \\
		&   observation noise [$\%$]   &  $2$ \\
		&   action noise [$\%$]  & $5$  \\
		&   L [m] &  $200$ \\
	
		\hline   
		Yaw Control
		& 	training time [day] & $1$\\
		& 	est. wall clock time [hr] & $1.7$ \\
		 
		&   good PID $(k_p,k_i,k_d)$ &  $(1,0,.5)$ \\
		&   poor PID $(k_p,k_i,k_d)$   &  $(1,0,0)$ \\
		&   ($\alpha$, $\beta$)  &  $(.5,.5)$ \\
		
		&   reward scale  &  $.1$ \\
		&   $T_{success}$ [s]  &  $5$ \\
		&   $\epsilon$ [$\%$] &  $10$ \\
		&   $w_{0:1}$  &  $(1, 1)$ \\
		&   $c$  &  $.1$ \\
	
		\hline
		Blimp Control 
		& 	training time [day] & $28$\\
		& 	est. wall clock time [hr] & $48$ \\ 
		
		&   $k_{thrust}^{pid}$  &  $(.7,.01,.05)$ \\
		&   $k_{\psi}^{pid}$  &  $(.3,.003,.0075)$ \\
		&   $k_{z}^{pid}$  &  $(2,.02,1.0)$ \\
		&   ($\alpha$, $\beta$)  &  $(.5,.5)$ \\
		&   reward scale  &  $.05$ \\
		&   $\epsilon$ [m]  &  $5$ \\
		&   $w_{0:3}$  &  $(100, .9, .1, .1)$ \\
		&   $i_{0:3}$  &  $(.6, .2, .1, .1)$ \\
		&   $j_{0:1}$  &  $(.5, 1)$ \\
		&   $k_0$  &  $1$ \\
		&   $c$  &  $(.4,.4,.1,.05)$ \\
		\hline
	\end{tabular}}
\caption{\textit{horizon }and $\gamma$ are relatively large due to the long response time of the blimp. In the blimp control task, altitude tracking reward weight, $i_0$, has greater value than other planar terms since altitude loss is not considered as part of the success reward. \textit{reward scale} is tuned to keep value loss within the range $[-1, 1]$ to reduce the side-effect from the gradient clip and stabilize training.\label{tab:exp_parameter}}
\end{table}
 

\section{Experimental Evaluation}
\label{sec:4_evaluation}

\subsection{Yaw Control Task}
\label{sec:yaw-control-task-evaluation}
\begin{figure}[h!]
	\centering
	\begin{subfigure}[b]{0.5\textwidth}
		\centering
		\includegraphics[width=\textwidth]{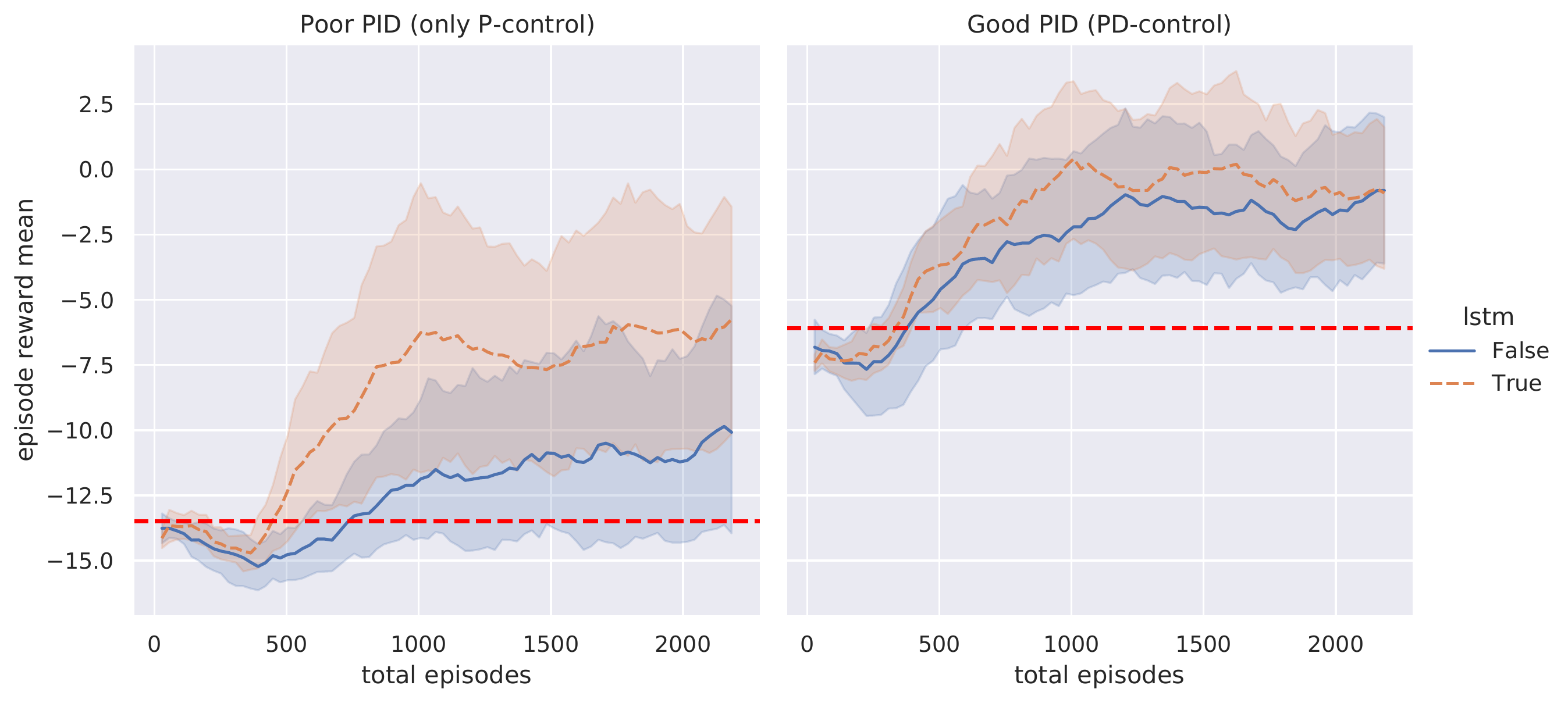}
		\caption{LSTM policy stabilizes and accelerates the training progress and reduces the performance drop.}
		\label{fig:yawcontrol_lstm_pid}
	\end{subfigure}
	
	\begin{subfigure}[b]{0.5\textwidth}
		\centering
		\includegraphics[width=\textwidth]{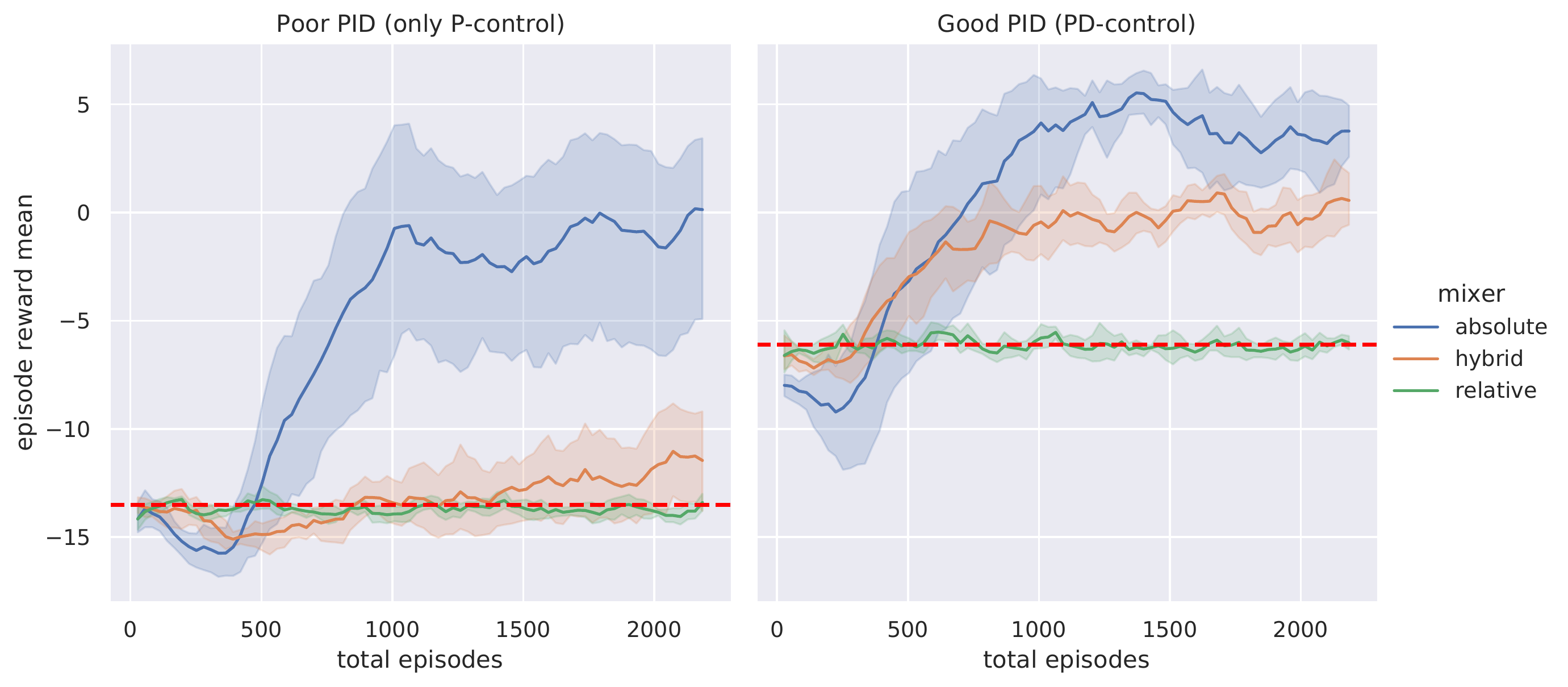}
		\caption{Absolute mixer is the most aggressive as it provides the most performance growth as well as performance drop during exploration, and vice versa.}
		\label{fig:yawcontrol_mixer_pid}
	\end{subfigure}
	\caption{Ablation study on the effect of LSTM and mixer type. Red dotted curve represent the average PID control performance.  Note that it is not a coincidence that the initial performance of the agent is similar to PID since the agent output layer weights are initialized close to zero.}
	\label{fig:yawcontrol_pid}
\end{figure}

Fig.~\ref{fig:yawcontrol_lstm_pid} shows that the final performance of DRRL with LSTM architecture increases the performance of good/poor PID control by 47\%/13\%. Without LSTM, the improvement is only 33\%/13\%. The maximum performance drop during exploration is 7\%/25\% with LSTM and 31\%/47\% without LSTM.
This result suggests that the LSTM is an important building block for our DRRL framework. It can effectively stabilize and accelerate the training progress and reduce the performance drop during training. 

The properties of the mixer type can be observed through the training progress in Fig.~\ref{fig:yawcontrol_mixer_pid}. Unsurprisingly, the absolute mixer has the largest performance growth and drop during training and achieves the highest final performance for both good and poor PIDs, since it grants the DRRL agent more control authority. On the contrary, the relative mixer neither improves nor degrades the PID performance. Equation (\ref{eq:relative_mix}) suggests that, as the agent control is dependent on the PID, the DRRL agent has no control authority if PID control is small. That is, when the yaw error is close to 0, P control and DRRL agent offer no control to reduce the angular velocity and overshoot the target angle. Consequently, the marginally stable system property remains unchanged. Lastly, the hybrid mix retains the properties from both mixers and achieves the intermediate performance as expected. Although absolute mixer appears to be the best design, it is important to note that it can reduce the training stability. This effect is amplified in higher dimensional space. 

We draw the following conclusions from the yaw control experiment: 1) The LSTM plays an important role to stabilize and accelerate training. 2) The choice of the mixer has significant impact on the training stability and final performance. 3) In addition to  the importance of LSTM and mixer, the design choice of the PID controller is also important. By including a derivative term boosts nearly 50\% of the initial performance. 

\subsection{Blimp Control Task}
\label{sec:blimp-control-task-evaluation}

\begin{figure}[t]
	\centering
	\includegraphics[width=0.35\textwidth]{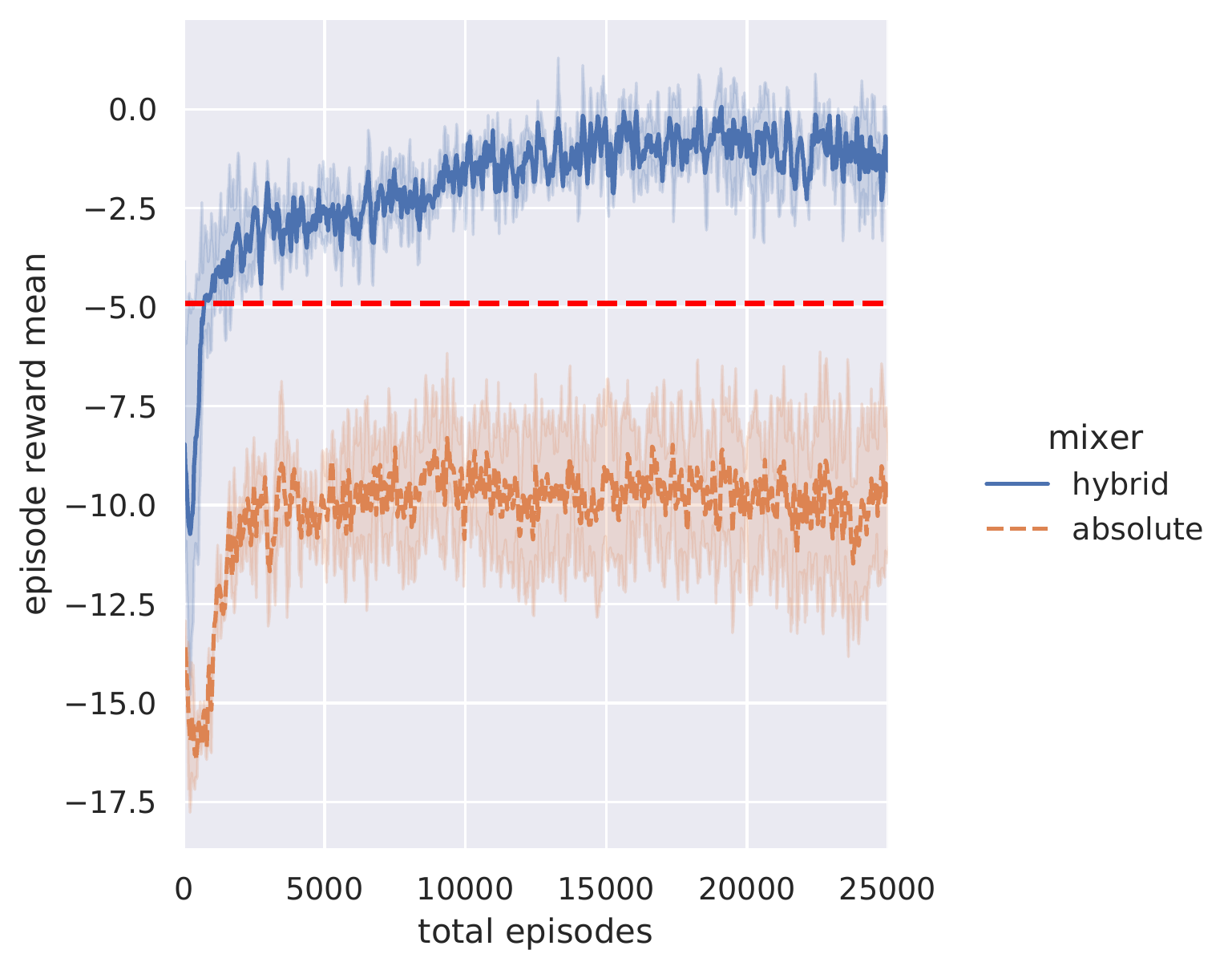}
	\caption{The blimp control task training progress with our DRRL integration. The red dotted line indicates the PID performance. The hybrid mixer provides better training stability compare to absolute mixer.}
	\label{fig:blimpcontrol_mixer}
\end{figure}

Following the conclusion from the ablation study (\ref{sec:yaw-control-task-evaluation}), we design the DRRL with LSTM policy and absolute mixer as well as with the hybrid mixer. Even though the absolute mixer appears to be the best configuration in simpler task, the agent with this mixer consistently failed to obtain any functional policy and got stuck in a local optima where the policy always commands maximum tail rotor and results in repeated rotation movement. The hybrid mixer, on the other hand, successfully stabilizes the training progress (Fig.~\ref{fig:blimpcontrol_mixer}). It reaches 60\% of the final performance within the first 2000 episodes and continues to grow steadily. 

\begin{figure}[h!]
	\centering
	\begin{subfigure}[b]{0.5\textwidth}
		\centering
		\includegraphics[width=\textwidth]{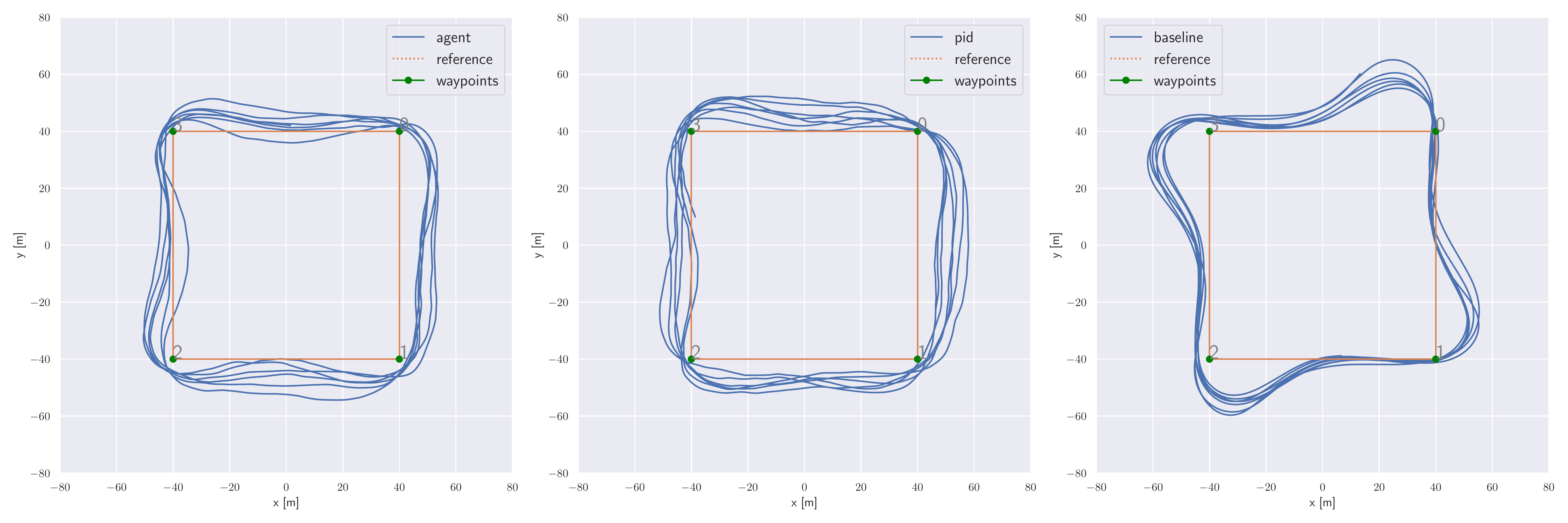}
		\caption{The square trajectory has long edge and sharp corner. It is served to test tracking performance and overshoot reduction.}
		\label{fig:square}
	\end{subfigure}
	
	\begin{subfigure}[b]{0.5\textwidth}
		\centering
		\includegraphics[width=\textwidth]{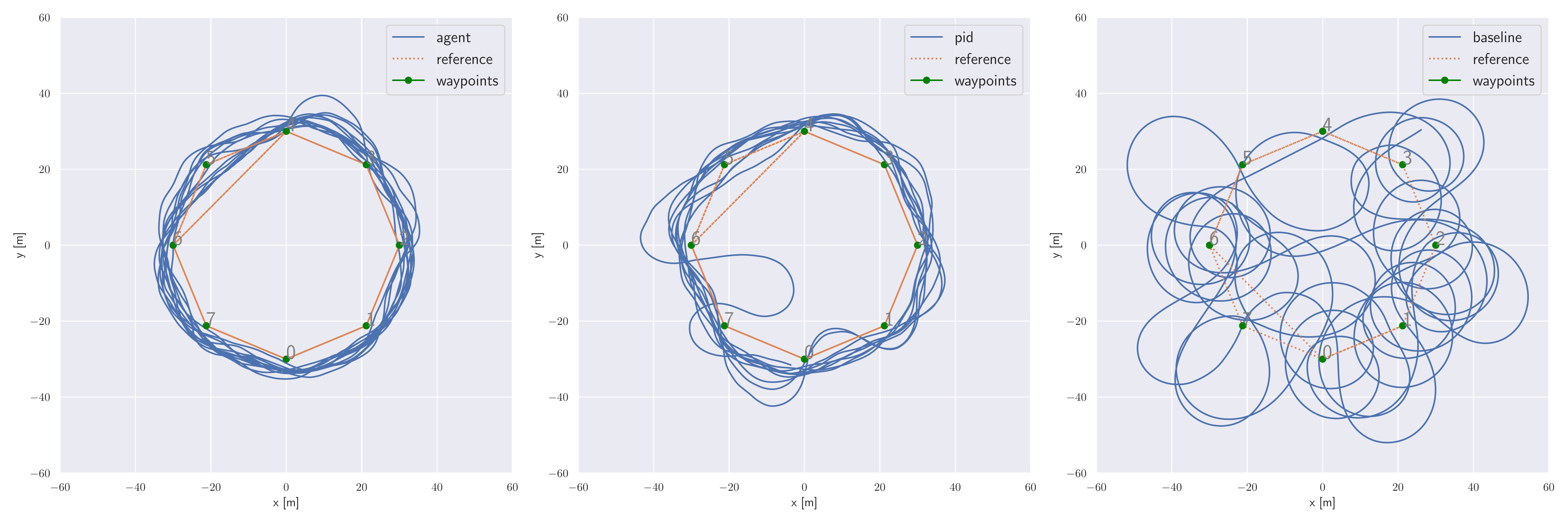}
		\caption{The coil trajectory has shorter distance between the waypoints. It is served for agility and altitude control test. }
		\label{fig:coil}
	\end{subfigure}
	\caption{Behavior comparison of different controllers in no wind condition.}
	\label{fig:trajectory}
\end{figure}

As demonstrated in Fig.~\ref{fig:trajectory}, the agent successfully tracks both the square and coil trajectory, which implies that it does not overfit to any specific tracks and can generalize to any desired trajectory in the 3D space. The baseline trajectory is more consistent compare to the others. This is because EKF provides smoother state estimation while both the DRRL agent and PID control receive only noisy raw observations. On the other hand, although the trajectories of the DRRL agent and the PID controller look fairly similar, the DRRL agent has much less overshoots compared to both the other methods in the coil trajectory (Fig.~\ref{fig:coil}) as it can observe the subsequent waypoint. In the coil trajectory, we observe that the baseline struggles following denser waypoints. To prevent altitude loss, the baseline applies a constraint on the maximum yaw angular rate, which limits the maximum turn radius and reduces its agility. 

In Tab.~\ref{tab:evaluation} we summarize the robustness tests and comparison of methods over different trajectory types, wind speeds and buoyancy. The DRRL agent receives highest amount of `total reward' in 4 out of 6 experiment combinations. Higher `success reward' implies that the agent can track more waypoints within the total time span. During experiments, although the desired velocity is $3$m/s, the baseline seems to achieve only $2m/s$. As a consequences, it traverses less total distance and receives less amount of success reward. In terms of tracking reward, the baseline significantly outperform others. Since the tracking reward is dominated by the altitude loss, this suggests that the baseline can keep track of the altitude better than PID and DRRL agent. In the coil trajectory, although the PID and the DRRL agent can follow the trajectory well, we observe significant loss in altitude. PID control does not have sufficient speed to maintain the altitude and continues to sink, while the DRRL agent relies on the thrust vectoring to loiter at the desired altitude. Similarly, reducing the buoyancy can impair the altitude control of the RL agent and PID control. The baseline, while being worst at overall waypoint tracking, tracks the altitude well. It achieves this via thrust vectoring and because, by design, the baseline's primary task includes maintaining the altitude.

\begin{table}[h!]
	\centering
	\resizebox{0.5\textwidth}{!}{%
		\begin{tabular}{ |p{1.3cm}|p{1.7cm}|p{1.5cm}||p{1.5cm}|p{1.5cm}|p{1.5cm}|p{1.5cm}|p{1.5cm}| }
			\hline
			\multicolumn{8}{|c|}{Robustness Evaluation} \\
			\hline
			Trajectory & wind[$m/s$]& controller & $r$ & $r^{success}$ & $r^{track}$ & $r^{act}$ & $r^{bonus}$ \\
			\hline
			square 
			& 0 & DRRL & ~0.0017 & 0.0027 & -0.2143 & -0.4308 & -0.0107 \\
			&  &PID  & -0.0055 & 0.0030 & -0.4073 & -0.4177 & -0.0112 \\
			&  & Baseline & ~0.0025 & 0.0020 & -0.1050 & -0.5896 & -0.0136 \\
			& 0.5 & DRRL  & ~0.0058 & 0.0036 & -0.2128 & -0.5106 & -0.0112  \\
			&  &PID  & -0.0050 & 0.0029 & -0.3901 & -0.4186 & -0.0111  \\
			&  & Baseline  & ~0.0020 & 0.0019 & -0.1101 & -0.5879 & -0.0128  \\
			& 1 &DRRL   & ~0.0022 & 0.0026 & -0.1927 & -0.4304 & -0.0111 \\
			&  &PID  & -0.0077 & 0.0022 & -0.3601 & -0.4437 & -0.0133\\
			&  & Baseline  & -0.0002 & 0.0016 & -0.1123 & -0.5849 &-0.0141\\
			\hline
			coil 
			& 0  &DRRL  & ~0.0189 & 0.0082 & -0.4253 & -0.6330 & -0.0100  \\
			&  &PID  & ~0.0282 & 0.0106 & -0.4890 & -0.5758 & -0.0118   \\
			&  & Baseline & -0.0032 & 0.0013 & -0.1501 & -0.5795 & -0.0207  \\			
			& 0.5 &DRRL   & ~0.0451 & 0.0118 & -0.2448 & -0.6172 & -0.0118\\
			&  &PID   & ~0.0363 & 0.0118 & -0.4435 & -0.5733 & -0.0124\\
			&  & Baseline  & ~0.0017 & 0.0021 & -0.1256 & -0.5786 & -0.0218 \\
			& 1 &DRRL   & ~0.0397 & 0.0108 & -0.2504 & -0.6211 & -0.0124 \\
			&  &PID   & ~0.0245 & 0.0089 & -0.3737 & -0.6074 & -0.0122  \\
			&  & Baseline & -0.0014 & 0.0016 & -0.1373 & -0.5780 & -0.0192  \\
			\hline
			\hline
			& buoyancy[\%]& &  &  &  &  &  \\
			\hline
			square
			& 0.93  &DRRL  & -0.0001 & 0.0026 & -0.2466 & -0.4369 & -0.0117  \\
			&   &PID  & -0.0009 & 0.0025 & -0.4232 & -0.4231 & -0.0113  \\
			&  &Baseline & ~0.0026 & 0.0019 & -0.1049 & -0.5899 & -0.0137 \\
			& 1.07  &DRRL  & ~0.0043 & 0.0030 & -0.1861 & -0.4262 & -0.0114  \\
			&   & PID  & -0.0037 & 0.0030 & -0.3707 & -0.4226 & -0.0113  \\
			&  &Baseline & ~0.0022 & 0.0018 & -0.1044 & -0.5816 & -0.0135 \\

			\hline
			
	\end{tabular}}
	\caption{The wind has the same speed for all runs in one trial but the directions are uniformly sampled. Rewards presented are averaged over all timesteps and trials. Higher reward is preferred. The overall performance is indicated by the reward column, $r$.  \label{tab:evaluation}}
\end{table}

\begin{figure}[ht!]
	\centering
	\begin{subfigure}[b]{0.5\textwidth}
        \begin{tabular}{c}
        \includegraphics[width=\textwidth]{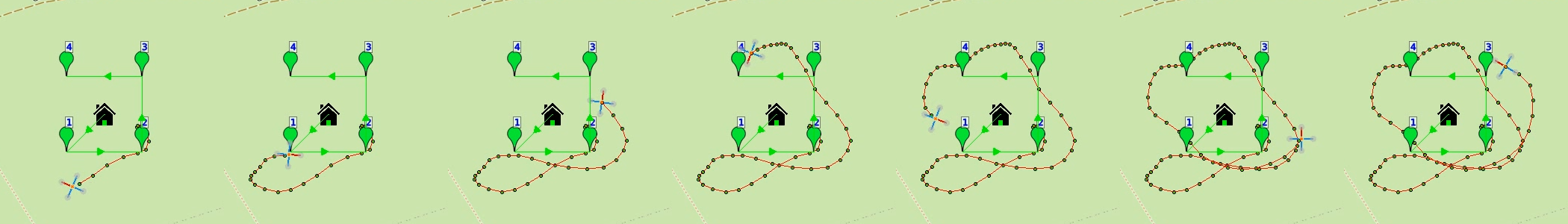}\\
        \vspace*{-5pt}  \includegraphics[trim={2cm 0.5cm 3cm 2cm},clip, width=\textwidth]{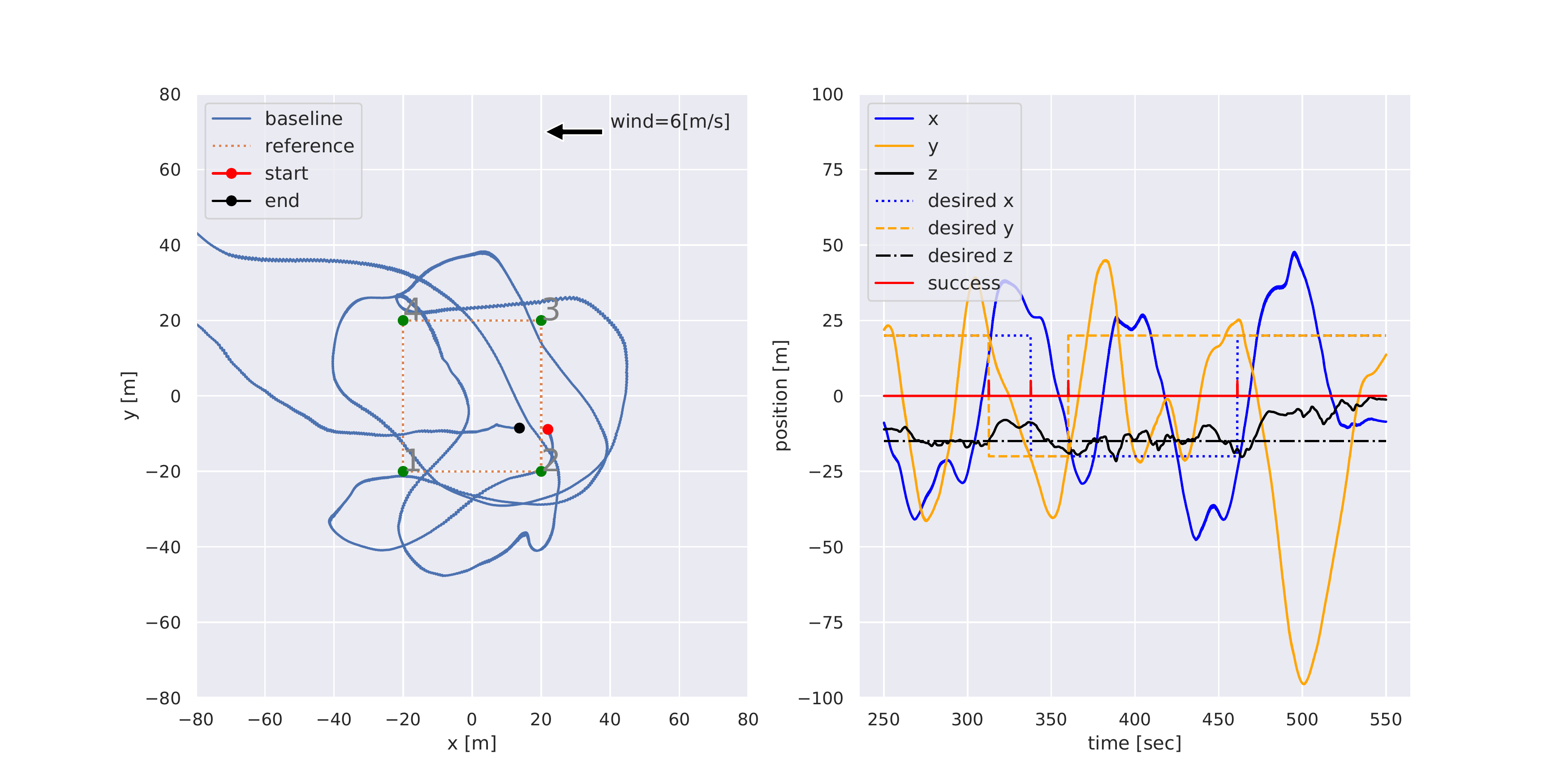} \\
        \end{tabular}
		\caption{Real world baseline flight.}
		\label{fig:PID_real}
	\end{subfigure}
	
	\vspace*{10pt}
	
	\begin{subfigure}[b]{0.5\textwidth}
        \begin{tabular}{c}
        \includegraphics[width=\textwidth]{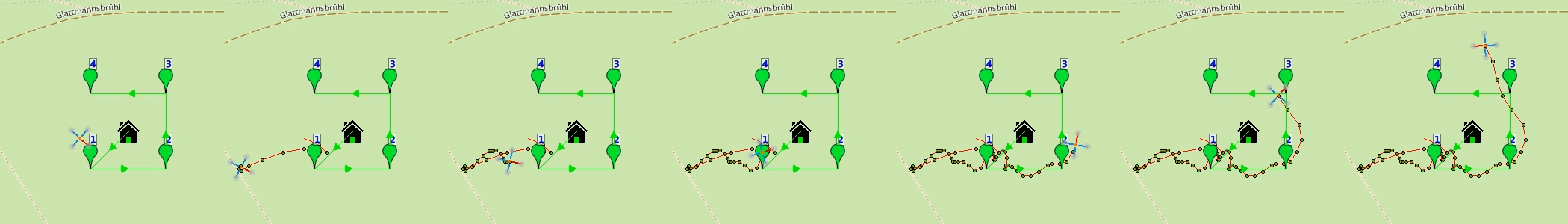}\\
        \vspace*{-5pt}  \includegraphics[trim={2cm 0.5cm 3cm 2cm},clip, width=\textwidth]{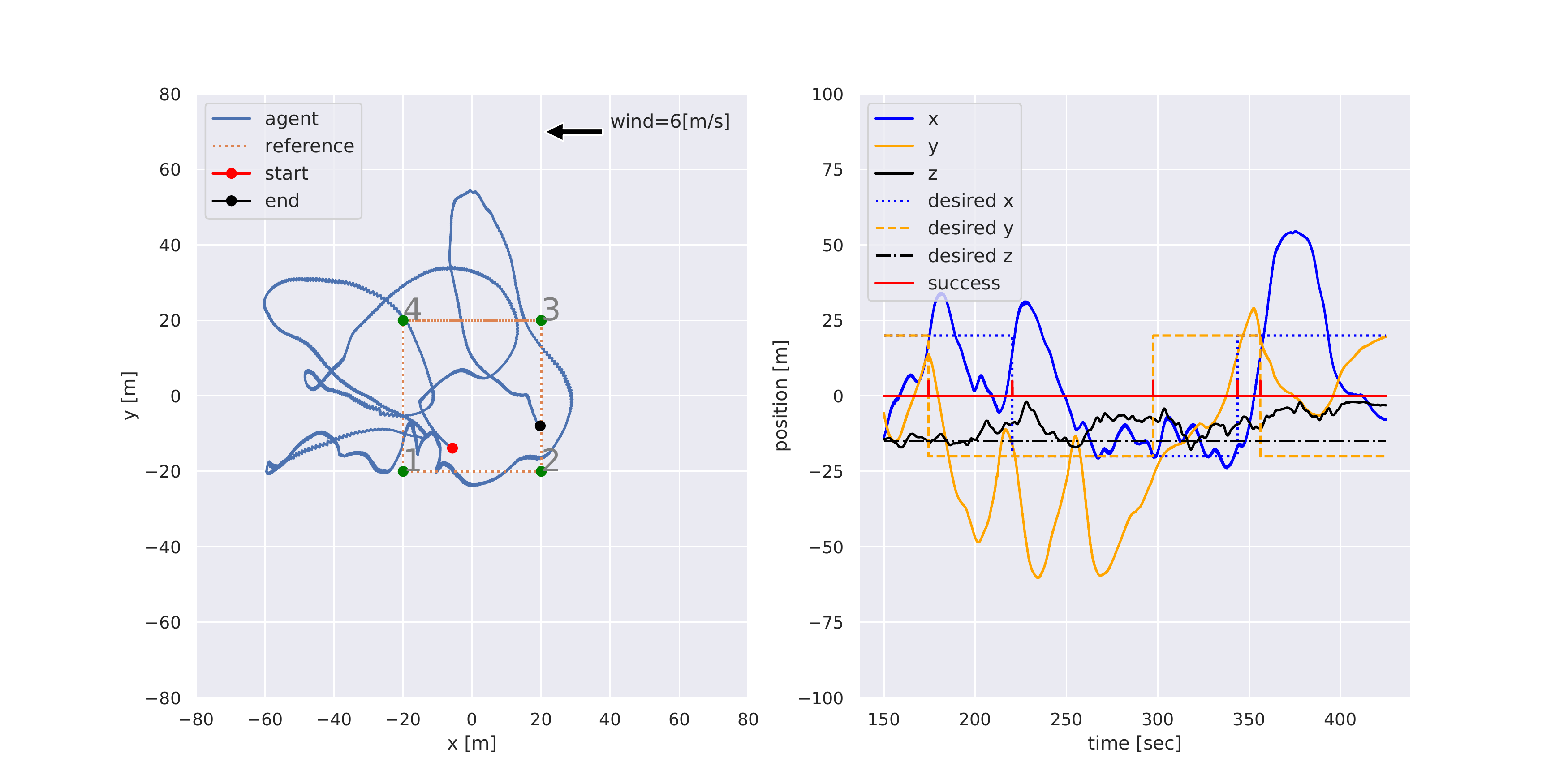} \\
        \end{tabular}
		\caption{Real world DRRL agent's flight.}
		\label{fig:Rl_real}
    \end{subfigure}
    \caption{Real World Experiments: Although the flight context is the same for both real-world flights, the wind gusts arrive at different times. Therefore, the baseline is only provided as a reference instead of a comparison. The top rows in each of the above figures show a part of the trajectory taken by the blimp to reach subsequent waypoints.}
    
    \label{fig:real_flight}
\end{figure}

\subsection{Real World Test}


The result of the real test flight is displayed in Fig.\ref{fig:real_flight} and Table\ref{tab:real-flight-evaluation}. We reduce the square size to $40$ meters as oppose to $80$ meter in simulation due to the limitation of the test field. The wind speed was measured in average $6m/s$ which was 4 times more than the DRRL agent had experienced in the simulation. Nevertheless, the DRRL agent could still hold its own position under the gusts and successfully reached several waypoints. Note that the row of trajectory snapshots in Fig.~\ref{fig:real_flight} show only a part of the complete trajectory. We also provide the baseline as a reference. But they are not comparable since the gusts were strong and arrived irregularly, and hence, the method that received more gusts would obtain less reward.




\begin{table}[h!]
	\centering
	\resizebox{0.5\textwidth}{!}{%
		\begin{tabular}{ |p{1.2cm}|p{1.3cm}||p{1.3cm}|p{1.3cm}|p{1.3cm}|p{1.3cm}|p{1.3cm}| }
			\hline
			\multicolumn{7}{|c|}{Real Flight Evaluation} \\
			\hline
			Trajectory & controller & $r$ & $r^{success}$ & $r^{track}$ & $r^{act}$ & $r^{bonus}$ \\
			\hline
			square 
			& DRRL & 0.0027 & 0.0022 & -0.0991 & -0.3969 & -0.0155 \\
			&Baseline  & 0.0006 & 0.0013 & -0.0961 & -0.3445 & -0.0109 \\
			\hline
			
	\end{tabular}}
	\caption{Averaged rewards by DRRL agent and baseline approaches during real-world flights.}
	\label{tab:real-flight-evaluation}
\end{table}


\section{Conclusions}

\label{sec:5_discussion}

In this work, we presented a novel framework based on DRRL for the blimp control task. It leverages an RL agent to improve the basic PID control performance through interaction with the environment. We presented and evaluated several techniques to stabilize the training progress and enhance the robustness of the trained RL agent, e.g., domain randomization, LSTM layer, and a hybrid mixer in the DRRL framework. Extensive robustness tests were conducted that demonstrated the DRRL agent's capability to improve the PID performance and outperform it as well as another baseline approach. Through real blimp flights in outdoor environment and windy conditions, we demonstrated that the trained policy could even generalize to a real scenario without any modification.

%
%
%

\bibliographystyle{IEEEtran}
\bibliography{biblio}

\end{document}